\documentclass[runningheads]{llncs}
\usepackage{graphicx}
\usepackage{times}
\usepackage{epsfig}
\usepackage{amsmath}
\usepackage{amssymb}
\usepackage{booktabs} 
\usepackage{enumerate}
\usepackage{adjustbox}
\usepackage{indentfirst}
\usepackage{array}
\usepackage{multirow}
\usepackage{float}
\usepackage{hhline}
\usepackage{bm}
\usepackage{subfigure}
\usepackage{color}
\usepackage{xcolor}

\begin{document}

\title{Visual Question Answering using Deep Learning: A Survey and Performance Analysis}

\author{Yash Srivastava \and
        Vaishnav Murali \and  
        Shiv Ram Dubey \and
        Snehasis Mukherjee
}


\institute{Computer Vision Group, \\
Indian Institute of Information Technology, Sri City, Chittoor, Andhra Pradesh, India.\\
\email{\{srivastava.y15, murali.v15, srdubey, snehasis.mukherjee\}@iiits.in} }

\maketitle

\begin{abstract}
The Visual Question Answering (VQA) task combines challenges for processing data with both Visual and Linguistic processing, to answer basic `common sense' questions about given images. Given an image and a question in natural language, the VQA system tries to find the correct answer to it using visual elements of the image and inference gathered from textual questions. In this survey, we cover and discuss the recent datasets released in the VQA domain dealing with various types of question-formats and robustness of the machine-learning models. Next, we discuss about new deep learning models that have shown promising results over the VQA datasets. At the end, we present and discuss some of the results computed by us over the vanilla VQA model, Stacked Attention Network and the VQA Challenge 2017 winner model. We also provide the detailed analysis along with the challenges and future research directions.\footnote{This paper is accepted in Fifth IAPR International Conference on Computer Vision and Image Processing (CVIP), 2020.}
\keywords{Visual Question Answering \and Artificial Intelligence \and Human Computer Interaction \and Deep Learning \and CNN \and LSTM.}
\end{abstract}

\section{Introduction}
\label{introduction}
Visual Question Answering (VQA) refers to a challenging task which lies at the intersection of image understanding and language processing. The VQA task has witnessed a significant progress the recent years by the machine intelligence community. The aim of VQA is to develop a system to answer specific questions about an input image. The answer could be in any of the following forms: a word, a phrase, binary answer, multiple choice answer, or a fill in the blank answer. Agarwal et al. \cite{vqa} presented a novel way of combining computer vision and natural language processing concepts of to achieve \textbf{Visual Grounded Dialogue}, a system mimicking the human understanding of the environment with the use of visual observation and language understanding.

\begin{figure}[!ht]
\centering
\includegraphics[width=\textwidth]{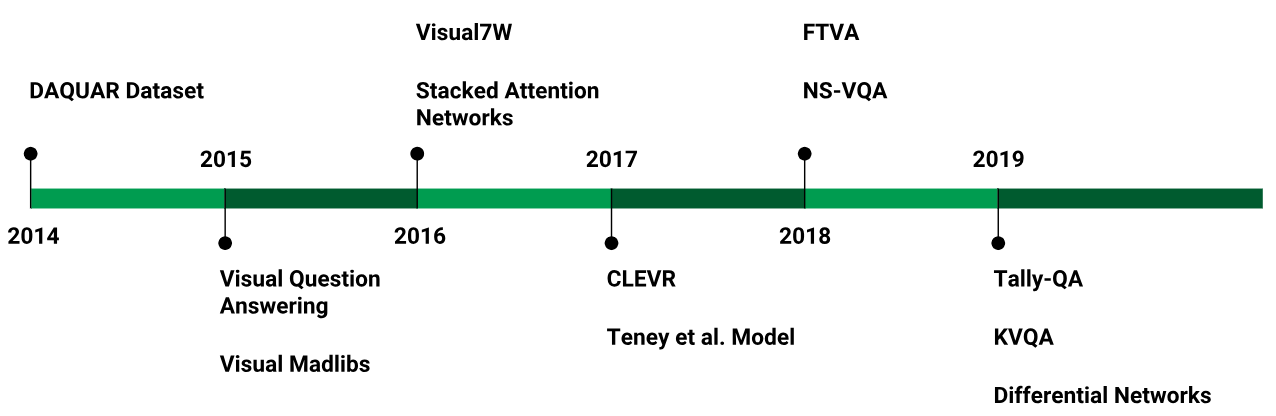}
\caption{Major Breakthrough Timeline in Visual Question Answering.}
\label{tbl:overview} 
\end{figure}

The advancements in the field of deep learning have certainly helped to develop systems for the task of Image Question Answering. Krizhevsky et al \cite{alexnet} proposed the AlexNet model, which created a revolution in the computer vision domain. The paper introduced the concept of Convolution Neural Networks (CNN) to the mainstream computer vision application. Later many authors have worked on CNN, which has resulted in robust, deep learning models like VGGNet \cite{vgg}, Inception \cite{inception}, ResNet \cite{resnet}, and etc. Similarly, the recent advancements in natural language processing area based on deep learning have improved the text understanding performance as well. The first major algorithm in the context of text processing is considered to be the Recurrent Neural Networks (RNN) \cite{rnn} which introduced the concept of prior context for time series based data. This architecture helped the growth of machine text understanding which gave new boundaries to machine translation, text classification and contextual understanding. Another major breakthrough in the domain was the introduction of Long-Short Term Memory (LSTM) architecture \cite{lstm} which improvised over the RNN by introducing a context cell which stores the prior relevant information. 

The vanilla VQA model \cite{vqa} used a combination of VGGNet \cite{vgg} and LSTM \cite{lstm}. This model has been revised over the years, employing newer architectures and mathematical formulations as seen in Fig. \ref{tbl:overview}. Along with this, many authors have worked on producing datasets for eliminating bias, strengthening the performance of the model by robust question-answer pairs which try to cover the various types of questions, testing the visual and language understanding of the system. Among the recent developments in the topic of VQA, Li et al. have used the context-aware knowledge aggregation to improve the VQA performance \cite{li2020boosting}. Yu et al. have perfomed the cross-modal knowledge reasoning in the network for obtaining a knowledge-driven VQA \cite{yu2020cross}. Chen et al. have improved the robustness of VQA approach by synthesizing the Counterfactual samples for training \cite{chen2020counterfactual}. Li et al. have employed the attention based mechanism through transfer learning alongwith a cross-modal gating approach to improve the VQA performance \cite{li2020visual}. Huang et al. \cite{huang2020aligned} have utilized the graph based convolutional network to increase the encoding relational informatoin for VQA. The VQA has been also observed in other domains, such as VQA for remote sensing data \cite{lobry2020rsvqa} and medical VQA \cite{zhan2020medical}.

In this survey, first we cover major datasets published for validating the Visual Question Answering task, such as VQA dataset \cite{vqa}, DAQUAR \cite{daquar}, Visual7W \cite{visual7w} and most recent datasets up to 2019 include Tally-QA \cite{tallyqa} and KVQA \cite{kvqa}. Next, we discuss the state-of-the-art architectures designed for the task of Visual Question Answering such as Vanilla VQA \cite{vqa}, Stacked Attention Networks \cite{san} and Pythia v1.0 \cite{pythia}. Next we present some of our computed results over the three architectures: vanilla VQA model \cite{vqa}, Stacked Attention Network (SAN) \cite{san} and Teney et al. model \cite{teney}. Finally, we discuss the observations and future directions. 

\begin{table*}[!t]
\centering
\caption{Overview of VQA datasets described in this paper.}
\centering
\begin{tabular}{|m{17mm}|m{10mm}|m{12mm}|m{35mm}|m{9mm}|m{20mm}|m{11mm}|}
\hline
\textbf{Dataset} & \textbf{\# Images} & \textbf{\# Questions} & \textbf{Question Type(s)} & \textbf{Venue} & \textbf{Model(s)} & \textbf{Accuracy}\tabularnewline
\hline
\hline
DAQUAR \cite{daquar}	&1449	&12468	&Object Identitfication	&NIPS 2014 & AutoSeg \cite{autoseg} &13.75\%\tabularnewline
\hline
VQA \cite{vqa}	&204721	&614163	&Combining vision, language and common-sense	&ICCV 2015 & CNN + LSTM &54.06\%\tabularnewline
\hline
Visual Madlibs \cite{madlibs}	&10738	&360001	&Fill in the blanks	&ICCV 2015 & nCCA (bbox) &47.9\%\tabularnewline
\hline
Visual7W \cite{visual7w}	&47300	&2201154	&7Ws, locating objects	&CVPR 2016 & LSTM + Attention &55.6\%\tabularnewline
\hline
CLEVR \cite{clevr}	&100000	&853554	&Synthetic question generation using relations	&CVPR 2017 & CNN + LSTM + Spatial Relationship &93\%\tabularnewline
\hline
Tally-QA \cite{tallyqa}	&165000	&306907	&Counting objects on varying complexities &AAAI 2019 & RCN Network &71.8\%\tabularnewline
\hline
KVQA \cite{kvqa}	&24602	&183007	&Questions based on Knowledge Graphs	&AAAI 2019 & MemNet&59.2\%\tabularnewline
\hline
\end{tabular}
\label{table:datasets}
\end{table*}
 


\section{Datasets}
\label{datasets}
The major VQA datasets are summarized in Table \ref{table:datasets}. We present the datasets below.

\textbf{DAQUAR:}
DAQUAR stands for Dataset for Question Answering on Real World Images, released by Malinowski et al. \cite{daquar}. It was the first dataset released for the IQA task. The images are taken from NYU-Depth V2 dataset \cite{nyud2}. The dataset is small with a total of 1449 images. The question bank includes 12468 question-answer pairs with 2483 unique questions. The questions have been generated by human annotations and confined within 9 question templates using annotations of the NYU-Depth dataset.

\begin{figure*}[!t]
\centering
\includegraphics[width=\textwidth]{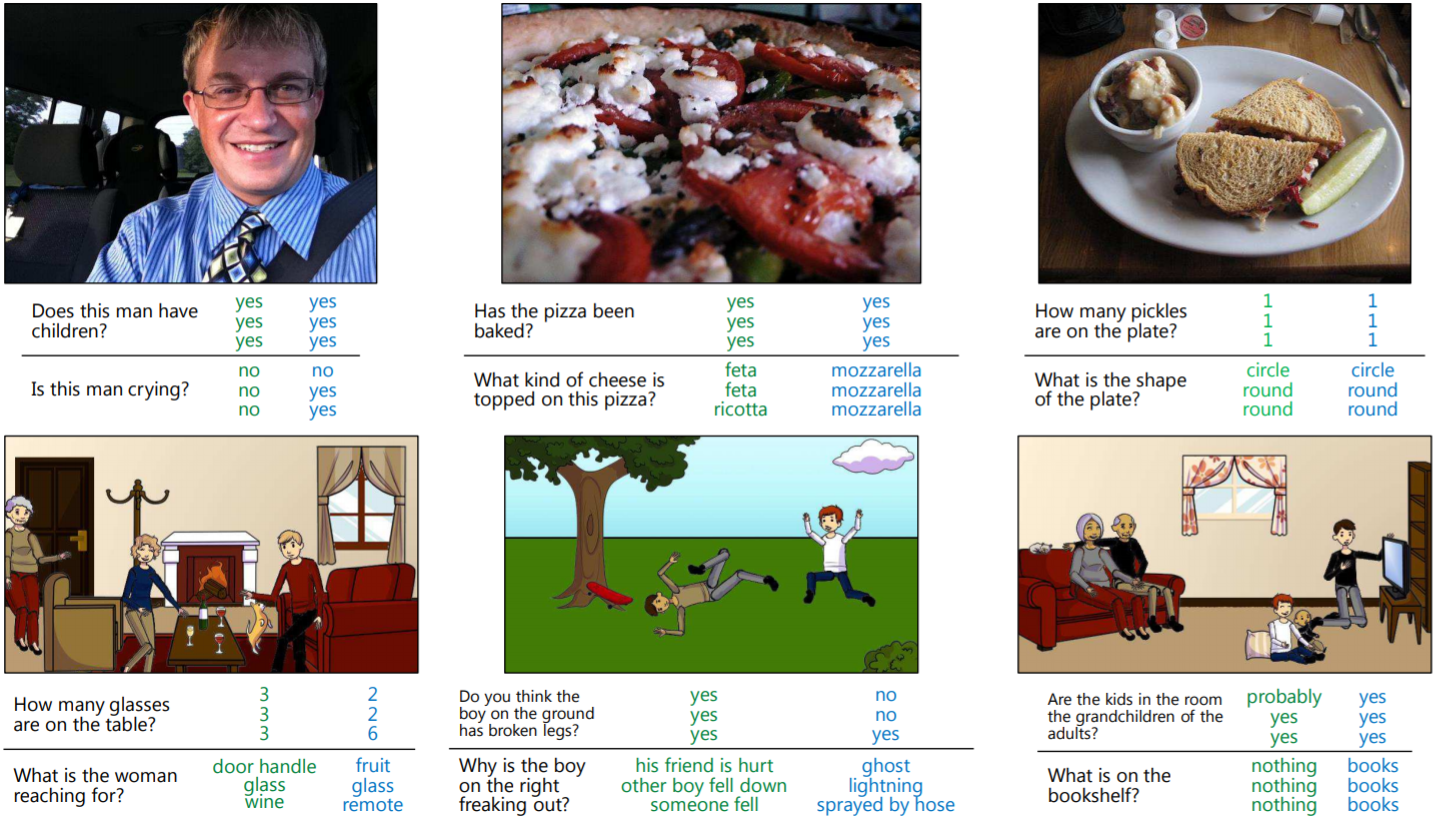}
\caption{Samples from VQA dataset \cite{vqa}.}
\label{dataset_vqa} 
\end{figure*}

\textbf{VQA Dataset:}
The Visual Question Answering (VQA) dataset \cite{vqa} is one of the largest datasets collected from the MS-COCO \cite{coco} dataset. The VQA dataset contains at least 3 questions per image with 10 answers per question. The dataset contains 614,163 questions in the form of open-ended and multiple choice. In multiple choice questions, the answers can be classified as: 1) Correct Answer, 2) Plausible Answer, 3) Popular Answers and 4) Random Answers. Recently, VQA V2 dataset \cite{vqa} is released with additional confusing images. The VQA sample images and questions are shown in Fig. \ref{dataset_vqa}.

\begin{figure*}[!t]
\centering
\includegraphics[width=1\textwidth]{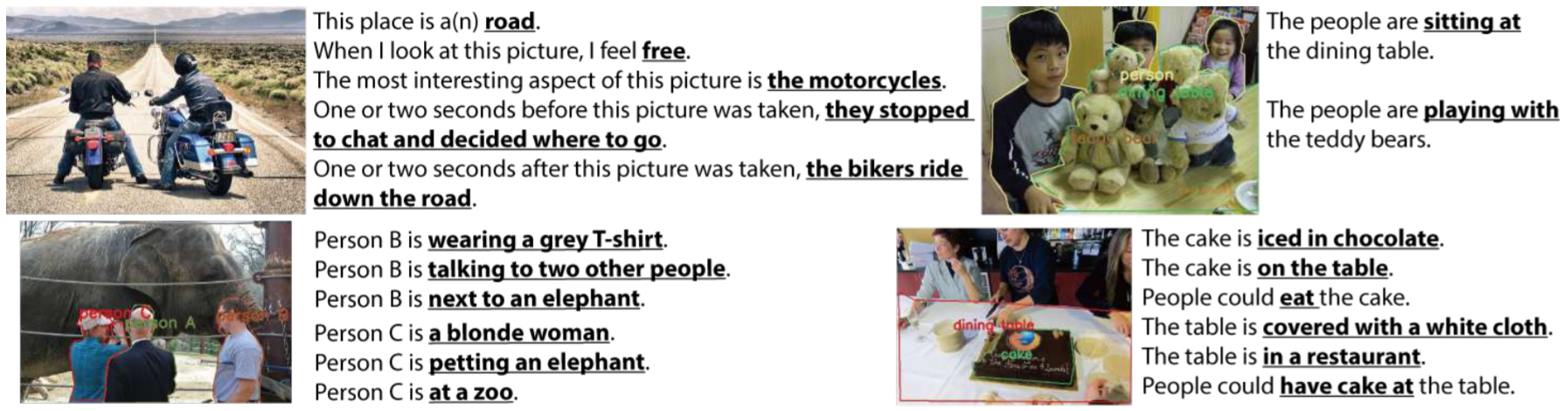}
\caption{Samples from Madlibs dataset \cite{madlibs}.}
\label{dataset_madlibs} 
\end{figure*}

\textbf{Visual Madlibs:}
The Visual Madlibs dataset \cite{madlibs} presents a different form of template for the Image Question Answering task. One of the forms is the fill in the blanks type, where the system needs to supplement the words to complete the sentence and it mostly targets people, objects, appearances, activities and interactions. The Visual Madlibs samples are shown in Fig. \ref{dataset_madlibs}.

\textbf{Visual7W:}
The Visual7W dataset \cite{visual7w} is also based on the MS-COCO dataset. It contains 47,300 COCO images with 327,939 question-answer pairs. The dataset also consists of 1,311,756 multiple choice questions and answers with 561,459 groundings.
The dataset mainly deals with seven forms of questions (from where it derives its name): What, Where, When, Who, Why, How, and Which. It is majorly formed by two types of questions. The ‘telling’ questions are the ones which are text-based, giving a sort of description. The ‘pointing’ questions are the ones that begin with ‘Which,’ and have to be correctly identified by the bounding boxes among the group of plausible answers.

\begin{figure*}[!t]
\centering
\includegraphics[width=1\textwidth]{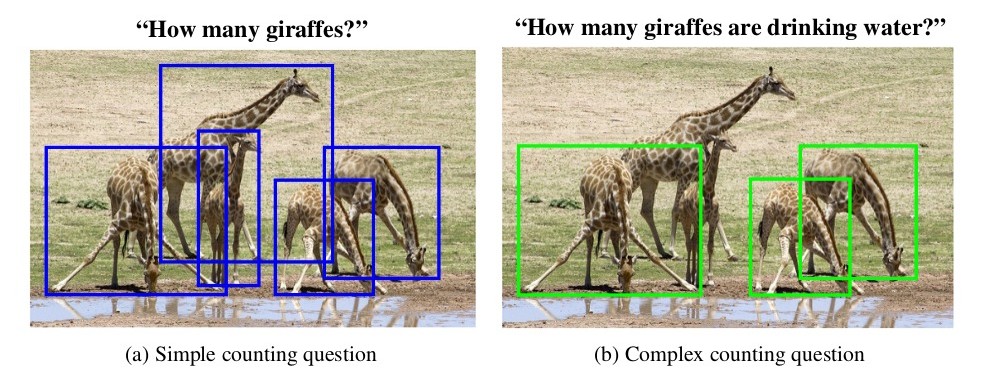}
\caption{Samples from Tally-QA dataset \cite{tallyqa}.}
\label{dataset_tally} 
\end{figure*}

\begin{figure*}[!t]
\centering
\includegraphics[width=1\textwidth]{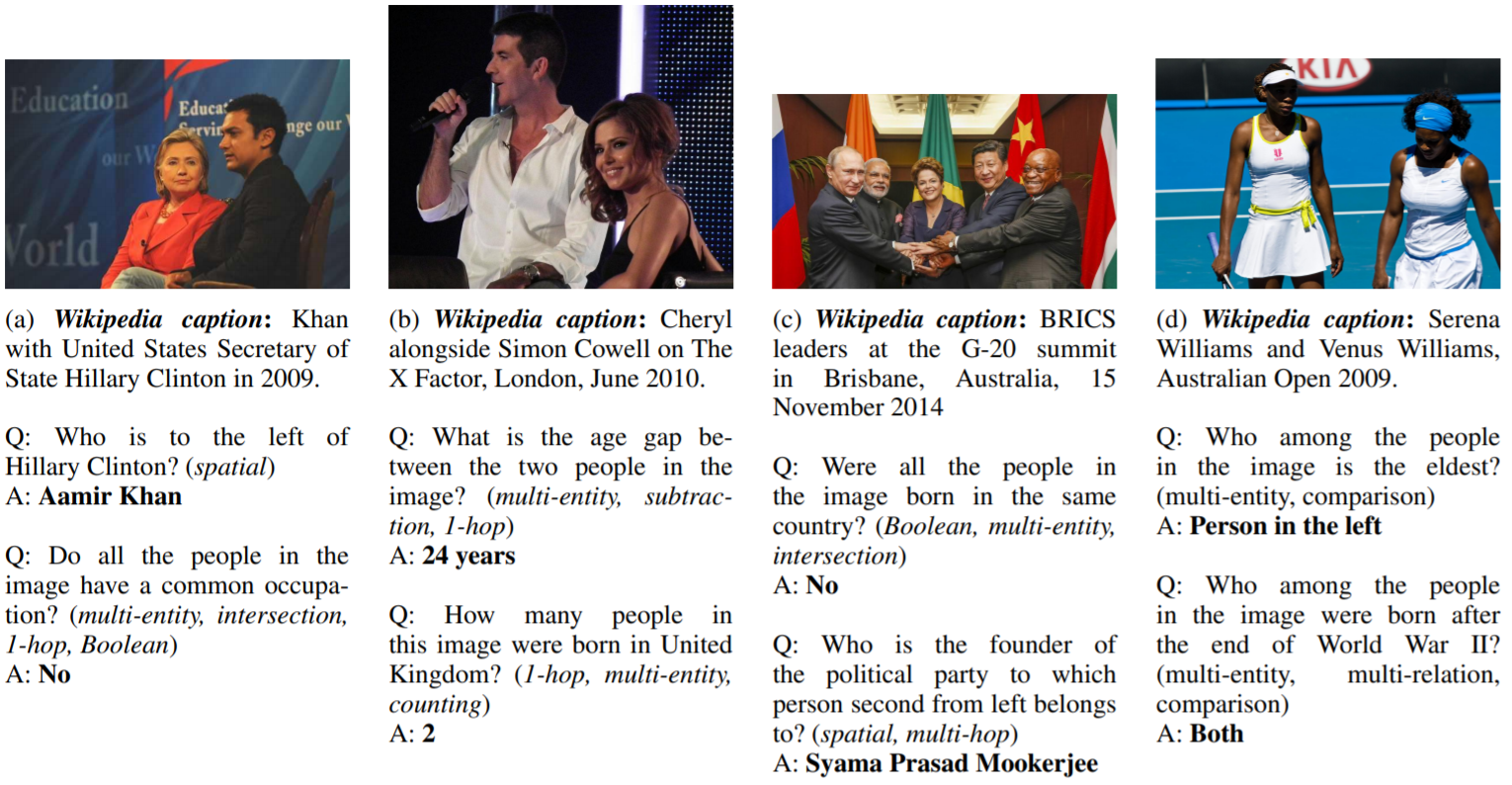}
\caption{Samples from KVQA dataset \cite{kvqa}.}
\label{dataset_kvqa} 
\end{figure*}

\textbf{CLEVR:}
CLEVR \cite{clevr} is a synthetic dataset to test the visual understanding of the VQA systems. The dataset is generated using three objects in each image, namely cylinder, sphere and cube. These objects are in two different sizes, two different materials and placed in eight different colors. The questions are also synthetically generated based on the objects placed in the image. The dataset also accompanies the ground-truth bounding boxes for each object in the image.

\textbf{Tally-QA:}
Very recently, in 2019, the Tally-QA \cite{tallyqa} dataset is proposed which is the largest dataset of object counting in the open-ended task. The dataset includes both simple and complex question types which can be seen in Fig. \ref{datasets}. The dataset is quite large in numbers as well as it is 2.5 times the VQA dataset. The dataset contains 287,907 questions, 165,000 images and 19,000 complex questions. The Tally-QA samples are shown in Fig. \ref{dataset_tally}.

\textbf{KVQA:}
The recent interest in common-sense questions has led to the development of Knowledge based VQA dataset \cite{kvqa}. The dataset contains questions targeting various categories of nouns and also require world knowledge to arrive at a solution. Questions in this dataset require multi-entity, multi-relation, and multi- hop reasoning over large Knowledge Graphs (KG) to arrive at an answer. The dataset contains 24,000 images with 183,100 question-answer pairs employing around 18K proper nouns. The KVQA samples are shown in Fig. \ref{dataset_kvqa}.









\begin{table*}[!t]
\caption{Overview of Models described in this paper. The Pythia v0.1 is the best performing model over VQA dataset.}
\centering
\begin{tabular}{|p{0.15\textwidth}|p{0.18\textwidth}|p{0.2\textwidth}|p{0.25\textwidth}|p{0.11\textwidth}|}
\hline
\textbf{Model} & \textbf{Dataset(s)} & \textbf{Method} & \textbf{Accuracy} & \textbf{Venue} \tabularnewline
\hline
\hline
Vanilla VQA \cite{vqa} & VQA \cite{vqa} & CNN + LSTM	&54.06 (VQA) & ICCV 2015\tabularnewline
\hline
Stacked Attention Networks \cite{san} & VQA \cite{vqa}, DAQAUR \cite{daquar}, COCO-QA \cite{cocoqa} & Multiple Attention Layers & 58.9 (VQA), 46.2 (DAQAUR), 61.6 (COCO-QA)	& CVPR 2016\tabularnewline
\hline
Teney et al. \cite{teney} & VQA \cite{vqa}	& Faster-RCNN 
+ Glove Vectors 
& 63.15 (VQA-v2) & CVPR 2018 \tabularnewline
\hline
Neural-Symbolic VQA \cite{nsvqa} & CLEVR \cite{clevr} & Symbolic Structure as Prior Knowledge & 99.8 (CLEVR) & NIPS 2018\tabularnewline
\hline
FVTA \cite{focal} & MemexQA \cite{memexqa}, MovieQA \cite{movieqa} & Attention over Sequential Data	& 66.9 (MemexQA), 37.3 (MovieQA) & CVPR 2018\tabularnewline
\hline
Pythia v1.0 \cite{py} & VQA \cite{vqa}	& Teney et al. \cite{teney} + Deep Layers & 72.27 (VQA-v2) & VQA Challenge 2018\tabularnewline
\hline
Differential Networks \cite{dn}	& VQA \cite{vqa}, TDIUC \cite{tdiuc}, COCO-QA \cite{cocoqa} & Faster-RCNN,
Differential Modules,
GRU 
& 68.59 (VQA-v2), 86.73 (TDIUC), 69.36 (COCO-QA) & AAAI 2019\tabularnewline
\hline
GNN \cite{zheng2019reasoning} & VisDial and VisDial-Q & Graph neural network & Recall: 48.95 (VisDial), 27.15 (VisDial-Q) & CVPR 2019\tabularnewline
\hline
\end{tabular}
\label{table:models}
\end{table*}

\section{Deep Learning Based VQA Methods}
\vspace{-0.2cm}
The emergence of deep-learning architectures have led to the development of the VQA systems. We discuss the state-of-the-art methods with an overview in Table \ref{table:models}.

\begin{figure}[!t]
\centering
\includegraphics[scale=5.0,  width=\textwidth]{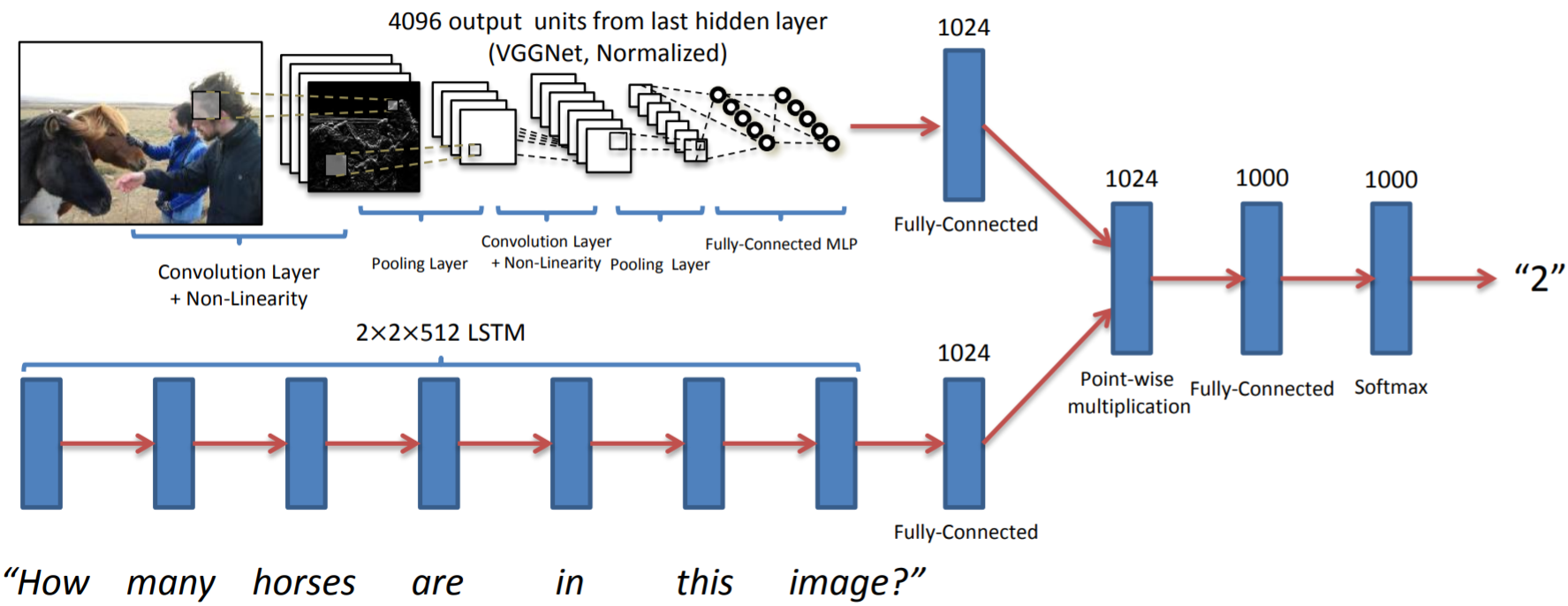}\\
\caption{Vanilla VQA Network Model \cite{vqa}.}
\label{fig:vqa} 
\end{figure}

\textbf{Vanilla VQA \cite{vqa}:}
Considered as a benchmark for deep learning methods, the vanilla VQA model uses CNN for feature extraction and LSTM or Recurrent networks for language processing. These features are combined using element-wise operations to a common feature, which is used to classify to one of the answers as shown in Fig. \ref{fig:vqa}.

\textbf{Stacked Attention Networks \cite{san}:}
This model introduced the attention using the softmax output of the intermediate question feature. The attention between the features are stacked which helps the model to focus on the important portion of the image.

\begin{figure}[!t]
\centering
\includegraphics[width=\textwidth]{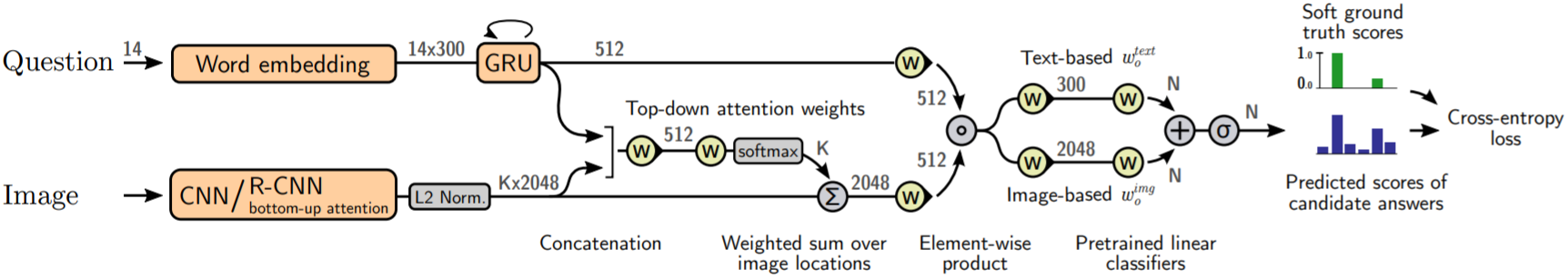}\\
\caption{Teney et al. VQA Model \cite{teney}}
\label{fig:teney} 
\end{figure}

\textbf{Teney et al. Model \cite{teney}:}
Teney et al. introduced the use of object detection on VQA models and won the VQA Challenge 2017. The model helps in narrowing down the features and apply better attention to images. The model employs the use of R-CNN architecture and showed significant performance in accuracy over other architectures. This model is depicted in Fig. \ref{fig:teney}.

\textbf{Neural-Symbolic VQA \cite{nsvqa}:}
Specifically made for CLEVR dataset, this model leverages the question formation and image generation strategy of CLEVR. The images are converted to structured features and the question features are converted to their original root question strategy. This feature is used to filter out the required answer.

\begin{figure}[!t]
\centering
\includegraphics[width=\textwidth]{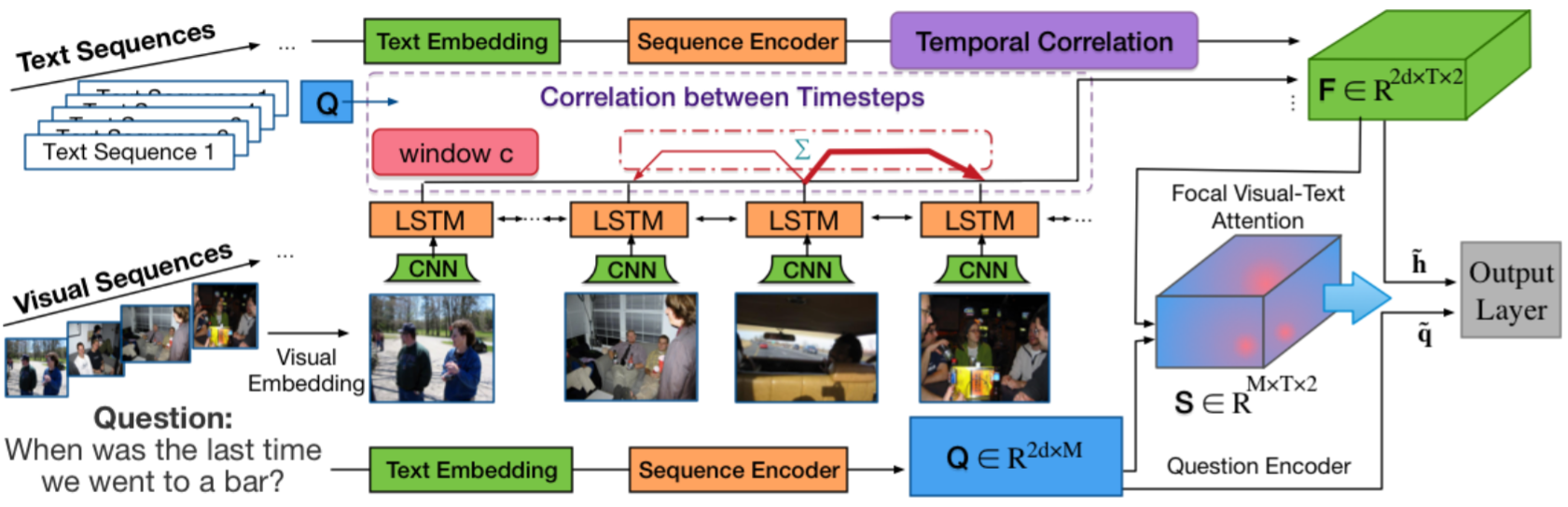}\\
\caption{Focal Visual Text Attention Model \cite{focal}}
\label{fig:focal} 
\end{figure}

\textbf{Focal Visual Text Attention (FVTA) \cite{focal}:}
This model combines the sequence of image features generated by the network, text features of the image (or probable answers) and the question. It applies the attention based on the both text components, and finally classifies the features to answer the question. This model is better suited for the VQA in videos which has more use cases than images. This model is shown in Fig. \ref{fig:focal}.

\begin{figure}[!t]
\centering
\includegraphics[width=\textwidth]{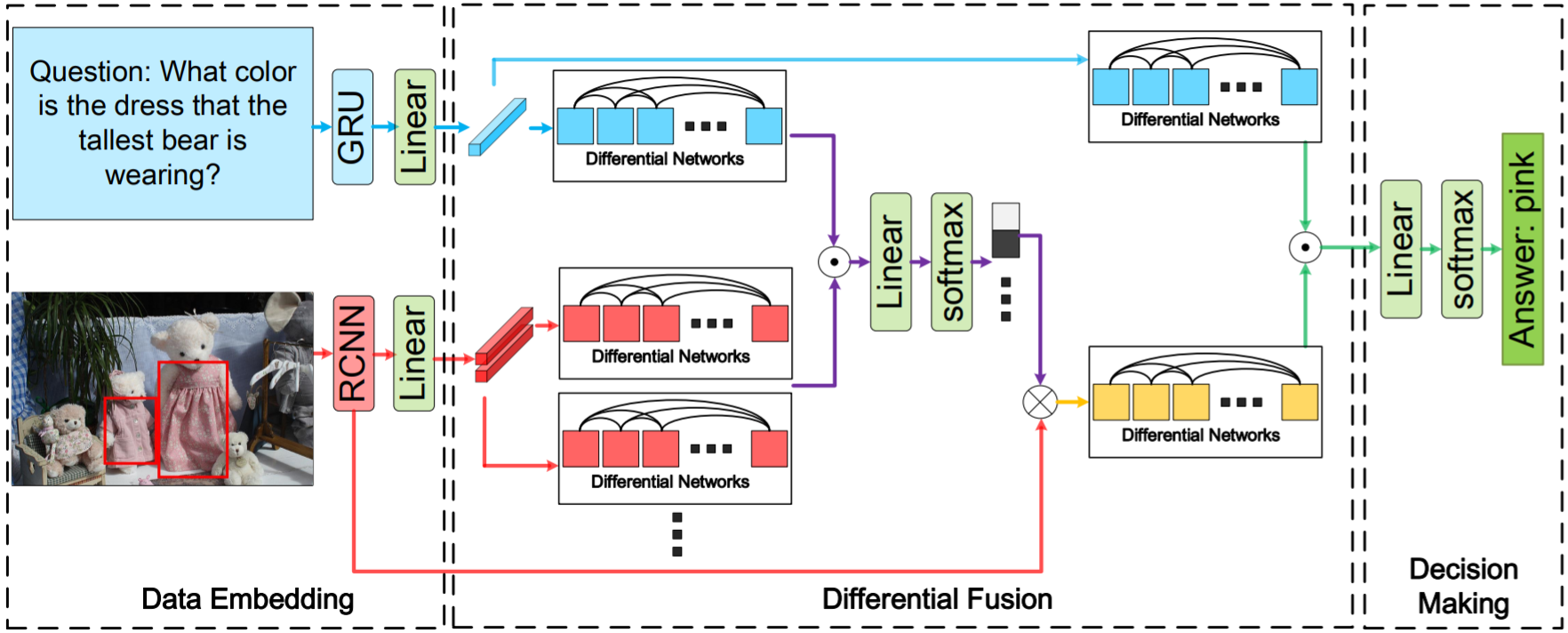}
\caption{Differential Networks Model \cite{dn}.}
\label{fig:dn} 
\end{figure}

\textbf{Pythia v1.0 \cite{py}:}
Pythia v1.0 is the award winning architecture for VQA Challenge 2018\footnote{\url{https://github.com/facebookresearch/pythia}}. The architecture is similar to Teney et al. \cite{teney} with reduced computations with element-wise multiplication, use of GloVe vectors \cite{pennington2014glove}, and ensemble of 30 models.

\textbf{Differential Networks \cite{dn}:}
This model uses the differences between forward propagation steps to reduce the noise and to learn the interdependency between features. Image features are extracted using Faster-RCNN \cite{frcnn}. The differential modules \cite{davqa} are used to refine the features in both text and images. GRU \cite{gru} is used for question feature extraction. Finally, it is combined with an attention module to classify the answers. The Differential Networks architecture is illustrated in Fig. \ref{fig:dn}.

\begin{figure}[!t]
\centering
\includegraphics[width=\textwidth]{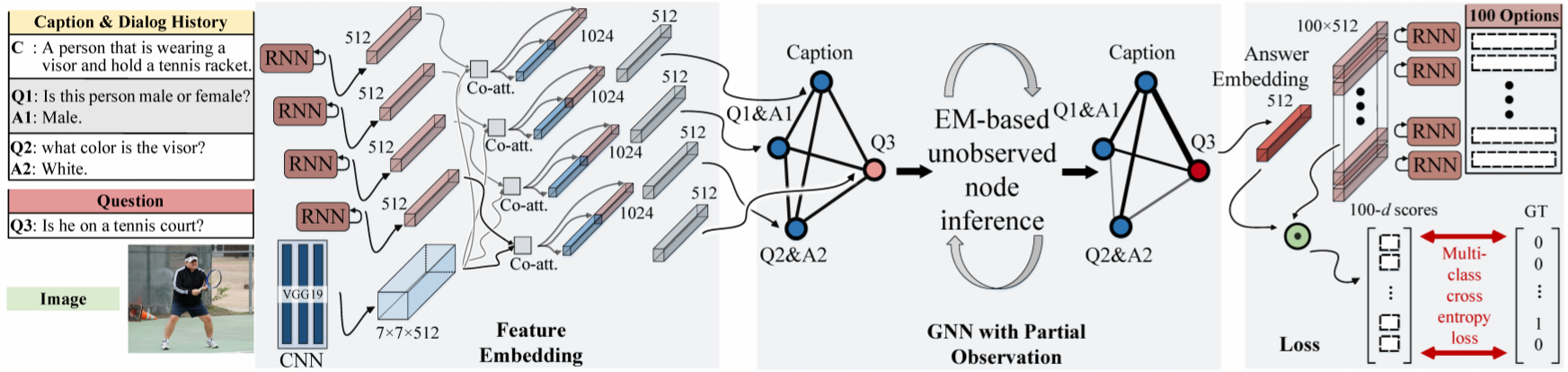}
\caption{Differentiable Graph Neural Network \cite{zheng2019reasoning}.}
\label{fig:gnn} 
\end{figure}

\textbf{Differentiable Graph Neural Network (GNN) \cite{zheng2019reasoning}:} Recently, Zheng at al. have discussed about a new way to model visual dialogs as structural graph and Markov Random Field.
They have considered the dialog entities as the observed nodes with answer as a node with missing value. This model is illustrated in Fig. \ref{fig:gnn}.




\section{Experimental Results and Analysis}
The reported results for different methods over different datasets are summarized in Table \ref{table:datasets} and Table \ref{table:models}. It can be observed that VQA dataset is very commonly used by different methods to test the performance. Other datasets like Visual7W, Tally-QA and KVQA are also very challenging and recent datasets. It can be also seen that the Pythia v1.0 is one of the recent methods performing very well over VQA dataset. The Differential Network is the very recent method proposed for VQA task and shows very promising performance over different datasets.

As part of this survey, we also implemented different methods over different datasets and performed the experiments. We considered the following three models for our experiments, 1) the baseline Vanilla VQA model \cite{vqa} which uses the VGG16 CNN architecture \cite{vgg} and LSTMs \cite{lstm}, 2) the Stacked Attention Networks \cite{san} architecture, and 3) the 2017 VQA challenge winner Teney et al. model \cite{teney}. We considered the widely adapted datasets such as standard VQA dataset \cite{vqa} and Visual7W dataset \cite{visual7w} for the experiments.
We used the Adam Optimizer for all models with Cross-Entropy loss function. Each model is trained for 100 epochs for each dataset.

\begin{table*}[!t]
\caption{The accuracies obtained using Vanilla VQA \cite{vqa}, Stacked Attention Networks \cite{san} and Teney et al. \cite{teney} models when trained on VQA \cite{vqa} and Visual7W \cite{visual7w} datasets.}
\centering
\begin{tabular}{|m{40mm}|m{25mm}|m{30mm}|}
\hline
\multirow{2}{*}{\textbf{Model Name}} & \multicolumn{2}{c|}{\textbf{Accuracy}}\\
\cline{2-3}
& \textbf{VQA Dataset} & \textbf{Visual7W Dataset} \tabularnewline
\hline
CNN + LSTM & 58.11 & 56.93 \tabularnewline
\hline
Stacked Attention Networks & 60.49 & 61.67\tabularnewline
\hline
Teney et al. & 67.23 & 65.82
 \tabularnewline
\hline
\end{tabular}
\label{table:observations}
\end{table*}


The experimental results are presented in Table \ref{table:observations} in terms of the accuracy for three models over two datasets. In the experiments, we found that the Teney et al. \cite{teney} is the best performing model on both VQA and Visual7W Dataset. The accuracies obtained over the Teney et al. model are 67.23\% and 65.82\% over VQA and Visual7W datasets for the open-ended question-answering task, respectively. The above results re-affirmed that the Teney et al. model is the best performing model till 2018 which has been pushed by Pythia v1.0 \cite{pythia}, recently, where they have utilized the same model with more layers to boost the performance. The accuracy for VQA is quite low due to the nature of this problem. VQA is one of the hard problems of computer vision, where the network has to understand the semantics of images, questions and relation in feature space.

\section{Conclusion}
The Visual Question Answering has recently witnessed a great interest and development by the group of researchers and scientists from all around the world. The recent trends are observed in the area of developing more and more real life looking datasets by incorporating the real world type questions and answers. The recent trends are also seen in the area of development of sophisticated deep learning models by better utilizing the visual cues as well as textual cues by different means. The performance of the best model is still lagging and around 60-70\% only. Thus, it is still an open problem to develop better deep learning models as well as more challenging datasets for VQA. Different strategies like object level details, segmentation masks, deeper models, sentiment of the question, etc. can be considered to develop the next generation VQA models.

\bibliographystyle{spmpsci}
\bibliography{Reference}

\end{document}